# Fully Convolutional Grasp Detection Network with Oriented Anchor Box

Xinwen Zhou, Xuguang Lan, Hanbo Zhang, Zhiqiang Tian, Yang Zhang and Nanning Zheng

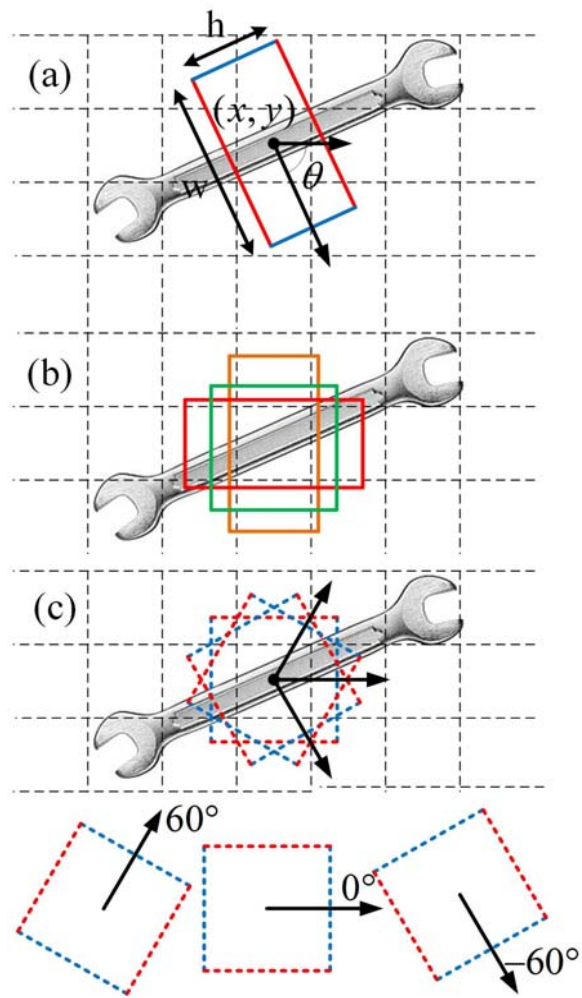

Fig. 1. (a) An example of five-dimensional representation of the grasp for a wrench.
(b) Non-oriented reference rectangles with 1 scale and 3 aspect ratios. Aspect ratio for red, green and orange rectangle is 1:2, 1:1 and 2:1 respectively.
(c) Our oriented anchor boxes at a grid cell. The orientations of these oriented anchor boxes is 60°, 0° and -60°.

*Abstract*—In this paper, we present a real-time approach to predict multiple grasping poses for a parallel-plate robotic gripper using RGB images. A model with oriented anchor box mechanism is proposed and a new matching strategy is used during the training process. An end-to-end fully convolutional neural network is employed in our work. The network consists of two parts: the feature extractor and multi-grasp predictor. The feature extractor is a deep convolutional neural network. The multi-grasp predictor regresses grasp rectangles from predefined oriented rectangles, called oriented anchor boxes, and classifies the rectangles into graspable and ungraspable. On the standard Cornell Grasp Dataset, our model achieves an accuracy of 97.74% and 96.61% on image-wise split and object-wise split respectively, and outperforms the latest state-of-the-art approach by 1.74% on image-wise split and 0.51% on object-wise split.

## I. INTRODUCTION

Grasping is an important ability for robots in human-machine cooperation under household and industrial scenes. Although human can instinctively execute grasps in an accurate, stable and rapid way even under the constantly changing environment, grasping is still challenging for robots. To grasp an object, robots need to first find the location for grasp. Inappropriate grasp locations will result in unsteadiness during the manipulation of object. It is necessary to find a more accurate way to detect grasp location. Thus, we propose a new approach to detect grasp locations for parallel plate gripper in a more accurate and rapid way.

Previous algorithms [1] [2] generate grasp configuration using 3-D model, which are powerful. However, complete and accurate 3-D model is hard to acquire in real world. RGB-D image is more convenient but it provides limited and noisy information. Taking RGB image as input, deep learning has achieved great success in grasp detection with its strong ability to learn useful features from data [3].

Guo et al. [4] and Chu et al. [5] have successfully applied deep learning to grasp detection. Guo et al. [4] introduce reference rectangles, also known as anchor boxes without varying orientation, in grasp detection shown in Fig. 1(b). Reference rectangles are a set of rectangles overlaid on the image at different spatial locations. In Fig. 1(b), a set of reference rectangles with 1 scale and 3 aspect ratios are at a spatial location. Note that these rectangles are parallel to horizontal axis. A hybrid deep architecture is employed, which fuses the visual sensing and tactile sensing, to predict three outputs (graspable, bounding box and orientation). The orientation of the grasp rectangles is quantized and predicted by classification. Using the same reference rectangle mechanism, Chu et al. [5], preprinted on ArXiv in Feb 2018, combine the classifications of graspable and orientation in [4] into one classifying problem. This modification transforms grasp detection into a combination of bounding box detection

Xinwen Zhou and Xuguang Lan are with the Institute of Artificial Intelligence and Robotics, the National Engineering Laboratory for Visual Information Processing and Applications, School of Electronic and Information Engineering, Xi'an Jiaotong University, No.28 Xianning Road, Xi'an, Shaanxi, China. E-mail: benbenben@stu.xjtu.edu.cn, xglan@mail.xjtu.edu.cn.

and orientation classification problems. Note that in both algorithms, the prediction of orientation is decoupled from the prediction of bounding box, while orientation is a geometric attribute of grasp rectangle instead of the semantic attribute.

Orientation matters much more to robotic grasp detection [3]. In most cases, a feasible gripper orientation for a given location is limited to a small range and is closely relevant to the location. For this reason, we propose the **oriented anchor box mechanism** that assigns the reference rectangles with different default rotation angles. In other words, we tile a set of rectangles with different default orientations across the image. Fig. 1(c) demonstrates an example of proposed oriented anchor boxes centered at one grid cell with the same scale and three default values (60°, 0° and -60°). To predict accurate grasps for a parallel-plate robotic gripper, we design a new fully convolutional network with this oriented anchor box mechanism.

Because the size of Cornell Grasp Dataset [6] is limited, the network is prone to overfit during training process. Parameters introduced by using fully connected layer will aggravate overfitting. Therefore, we employ end-to-end fully convolutional neural network, which consists of two parts: the feature extractor and multi-grasp predictor. Besides, we also pre-train our feature extractor ResNet [7] and implement extensive data augmentation to expand Cornell Grasp Dataset to avoid overfitting. The features extracted by ResNet are then fed into two sibling convolutional layers to regress the grasp configuration and predict the graspability of the corresponding grasp configuration.

During training process, we propose a new matching strategy for grasp detection, which combines point metric [3] and orientation constrain. It is faster than original matching strategy used in previous work [4].

Our model achieves an accuracy of 97.74% on image-wise split and 96.61% on object-wise, which outperforms Chu et al. [5], the current state-of-the-art algorithm, by 1.74% on image-wise split and 0.51% on object-wise split. Our work has two main contributions:

1) Considering the specialty of grasp detection, we design an oriented anchor box mechanism to improve the accuracy of grasp detection.

2) An efficient matching strategy is proposed, which is faster than the one in previous work.

The rest of the paper is organized as follows: The related work is discussed in Section II. Formulation of grasp detection problem in Section III. Our proposed approach is illustrated in Section IV. Detailed experiments setup is represented in Section V. We present our results in Section VI, then conclude in VII.

## II. RELATED WORK

To find a good grasp, many past works [1] [2] utilize full 3-D model of the object. However, when the robot is interacting with a new environment, complete 3-D model is an unknown priori. In real world, it is more convenient to capture RGB images than the reconstruction of 3-D model. Saxena et al. [8] learn grasps directly from images.

Lenz et al. [3] demonstrate that a five-dimensional grasp configuration in 2-D can be projected to 3-D, which simplifies the seven-dimensional representation proposed by Jiang et al. [9]. Therefore, grasp detection can be transformed to the problem similar to object detection. They also first introduce neural network as a classifier in the sliding window detection framework, to predict the existence of a good grasp in a small patch of input image. On Cornell Grasp Dataset, this method gets an accuracy of 73.9% on image-wise split at a speed of 13.5 seconds per frame.

Redmon et al. [10] raise the accuracy by big percentages with a locally constrained prediction mechanism and deeper network (AlexNet [11]). The locally constrained prediction mechanism predicts a grasp for each spatial location of the image, The RGD images (replacing the blue channel with depth) pass through the network once to obtain directly the detection results. The Direct Regression model can only predict one grasp for an image. In MultiGrasp Detection model, the input image is divided into $N \times N$ grid cells. For each grid cell, the MultiGrasp Detection model regresses a grasp rectangle and predicts its graspability. On image-wise split of Cornell Grasp Dataset, Direct Regression model has an accuracy of 84.4% and MultiGrasp Detection model gets an accuracy of 88.0% for Top1 detection result. Both models run at a speed of 3 frames per second. The acceleration mainly results from the computation afforded by GPU. According to [10], Direct Regression model usually suffers from averaging effect. For example, the predicted grasp of Direct Regression model for a big plate will locate at the center of the plate rather than its edge. Direct Regression model is hard to generalize to the images containing multiple objects with multiple grasp locations.

The best model proposed by Kumra et al. [12] outperforms that of Redmon et al. [10] by 1.21%. They use ResNet-50 [7] to regress directly a grasp configuration from RGB-D input (D for depth information). They also try uni-modal input like RGB data. Their model gets an accuracy of 88.84% on RGB data. Model on RGB-D data has an accuracy of 89.21%. The use of RGB-D input in this work improves marginally the accuracy by 0.37% compared with RGB input. Taking the same RGD input as the Direct Regression model of Redmon et al. [10], Kumra et al. [12] improve the accuracy by 4.13%. It means that deeper network brings better performance. The multi-modal grasp predictor runs at a speed of 9.71 frames per second.

Using ZF-net [13], similar to AlexNet, Guo et al. [4] outperform MultiGrasp Detection model of Redmon et al. [10] by 5.2% on Cornell Grasp Dataset. In [4], a hybrid deep architecture is used to produce three outputs (graspable, non-oriented bounding box and orientation). The network is capable to take tactile sensing and visual sensing as input. Both [4] and [10] have locally constrained prediction mechanism, but Guo et al. [4] goes further to associate each grid cell with default reference rectangles of various scales and aspect ratios. On Pascal VOC dataset, Faster RCNN [14] proves that the accuracy can be improved by introducing different scales and aspect ratios of the anchor boxes. However, this method only works on object-wise split of Cornell Grasp Dataset. On image-wise split, the setting of 54×54 scale with 1:1 aspect ratio achieves the best accuracy.

The performance even degrades when more scales and aspect ratios are applied. However, in their former work [15], a similar model achieves the best accuracy with 1 scale and 3 aspect ratios on CMU grasp dataset [16]. This is noteworthy, because it reveals that we need to choose suitable anchor box settings for different datasets. Unlike prior works, the prediction of orientation in this work is treated as a classification problem.

In Feb 2018, Chu et al. [5] turn grasp configuration regression problem into a combination of region detection and orientation classification problem. Similar to [4], the classification labels contain discrete orientations, except that a non-grasp label is also added. By doing this, graspable and orientation outputs in [4] are combined into one output. Note that Chu et al. [5] use 14×14×9 (1764) default reference rectangles while Guo et al. [13] only use 6×6×9 (324) default reference rectangles. With an accuracy of 96% on image-wise split and 96.1% on object-wise split, their best model outperforms Guo et al. [4] by 2.8% and 7%, respectively. Replacing VGG-16 [17] with ResNet-50 only brings 0.5% increase of accuracy on image-wise split. It means that adding convolutional layers has little influence on accuracy of image-wise split after the network reaches a certain depth. While the accuracy on object-wise split can be boosted when the network goes deeper.

From the perspective of objection detection , the models of [4][10][12] are one-stage detectors, which directly predict classes and anchor offsets without requiring a second stage per-proposal classification operation [18]. For one-stage detector, proposal is the predicted bounding box. The model in [5] is a typical two-stage detector, which requires a second stage to predict a class and class-specific refinement from the cropped features of the intermediate feature maps. Compared with the one-stage detector, the two-stage detector achieves a higher accuracy at the cost of running at a lower speed

### III. PROBLEM FORMULATION

Lenz et al. [3] propose five-dimensional grasp representation, which is widely used by the other works on grasp detection [4] [5] [10] [12]. In our work, we also adopt this representation. The five-dimensional grasp is represented as follow:

$$g = \{x, y, h, w, \theta\} \quad (1)$$

where $(x, y)$ is the center of oriented rectangle, $h$ is the height and $w$ is the width. $\theta$ is the angle between the horizontal axis and the moving direction of plates during the execution of grasp. An example has been represented in Fig. 1(a).

### IV. PROPOSED APPROCH

#### A. Oriented Anchor Box Mechanism

We introduce default angle term into original anchor box [14] in consideration of the importance of rotation angle in grasp detection. The input image is divided into $N \times N$ grid cells, and each grid cell has $k$ associated oriented anchor boxes centered at this grid cell. For a grid cell, we can predict $k$ types of grasps with different angles. Each oriented anchor box has its default rotation value. If $k$ equals to 6, then the grid cell has 6 oriented anchor box with rotation angle of 75°, 45°, 15°, -15°, -45° and -75°. In Fig. 2, we have a grid cell associated with oriented anchor boxes.

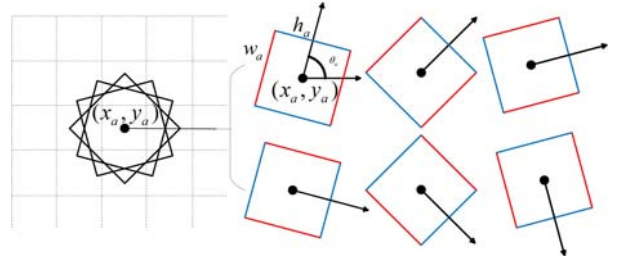

Fig. 2. Our oriented anchor box mechanism for a grid cell. In first line of the oriented anchor boxes, the rotation angle is 75°, 45° and 15°from left to right. The rotation angle of the oriented anchor boxes in the second line is -15°, -45° and -75°, from left to right.

$(x_a, y_a)$ are the center coordinates of the grid cell and its oriented anchor boxes. $w_a$ and $h_a$ are the predefined width and height of the oriented anchor box, and $\theta_a$ is the default rotation angle. Given the oriented anchor box configuration above and predicted grasp configurations, we employ following parameterization to project them into anchor offset:

$$\begin{aligned}
t_x &= (x - x_a)/w_a, & \hat{t}_x &= (\hat{x} - x_a)/w_a \\
t_y &= (y - y_a)/h_a, & \hat{t}_y &= (\hat{y} - y_a)/h_a \\
t_w &= \log(w/w_a), & \hat{t}_w &= \log(\hat{w}/w_a) \quad (2) \\
t_h &= \log(h/h_a), & \hat{t}_h &= \log(\hat{h}/h_a) \\
t_\theta &= (\theta - \theta_a)/(180/k), & \hat{t}_\theta &= (\hat{\theta} - \theta_a)/(180/k)
\end{aligned}$$

where $k$ is the number of the oriented anchor boxes associated with each gird cell. $x, y, w, h$ and $\theta$ denote the two coordinates of the predicted rectangle center, width, height and rotation angle, respectively. Variable $x, x_a$ and $\hat{x}$ are separately for predicted oriented rectangle, oriented anchor box and ground-truth grasp rectangle (likewise for $y, w, h, \theta$). The $t_x, t_y, t_w, t_h$ and $t_\theta$ are the offsets of the predicted rectangle. $\hat{t}_x, \hat{t}_y, \hat{t}_w, \hat{t}_h$ and $\hat{t}_\theta$ are the offsets of the ground-truth grasp rectangle with its matched oriented anchor box. We will discuss the detail of matching strategy in Section IV-C.

The size of the objects in dataset can vary greatly, which is the reason of introducing multiple scales in object detection. While this variation of size is less common in Cornell Grasp Dataset due to the constrained size of gripper and the fixed photographic viewpoint. In practice, we set the size of the oriented anchor boxes to be 54×54. It is greater than the size of grid cell so that the oriented anchor boxes can cover the image seamlessly. Our oriented anchor boxes vary in rotation angle rather than scale and aspect ratio. The scale, aspect ratio and the location of the oriented anchor box center will be refined by the process of regression. These oriented anchor boxes cover all the potential grasps of the input image in a sliding window fashion. How to choose a value for $k$ will be discussed in Section IV-C.

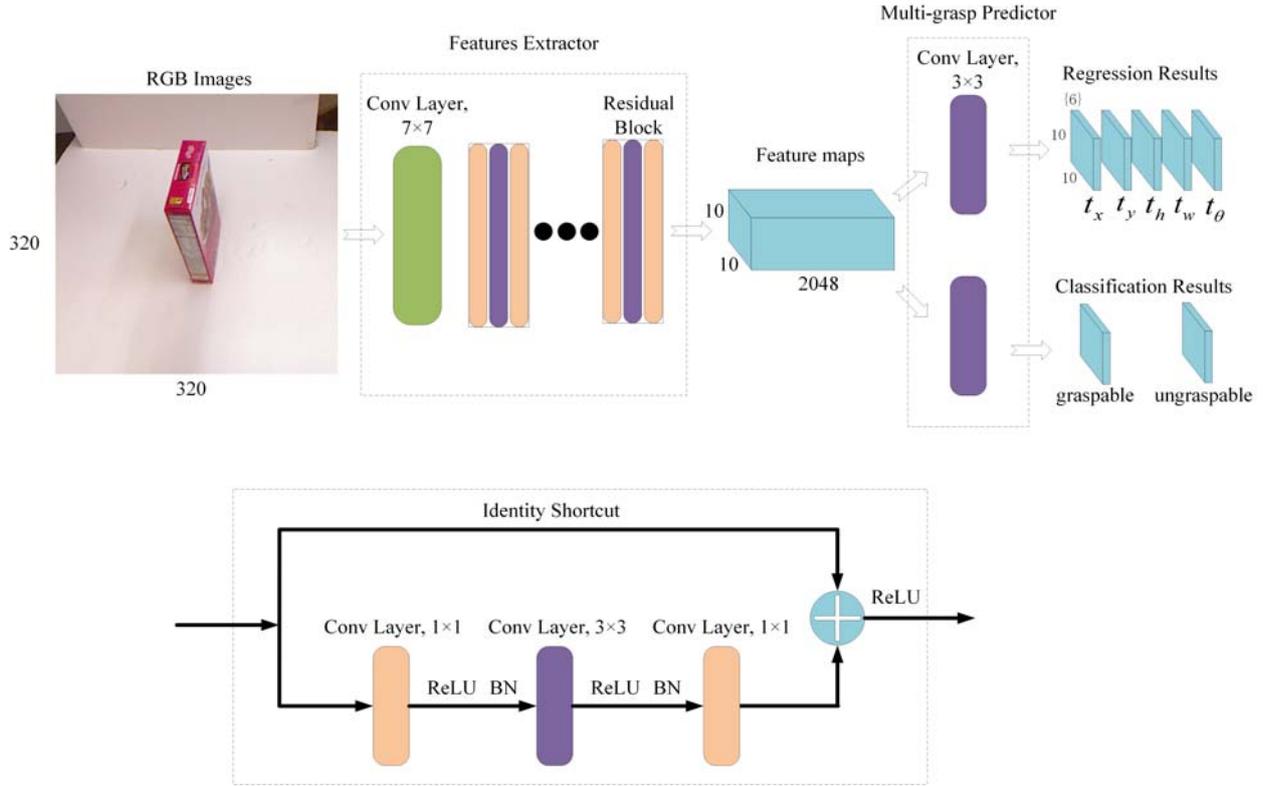

Fig. 3. Above: entire fully convolutional network. For ResNet-50, the number of residual blocks is 16. The number of residual blocks for ResNet-101 is 33. Below: the detailed architecture of residual block. Note: BN means Batch Normalization.

*B. Model Architecture*

Our model consists of a feature extractor and a multi-grasp predictor. The architecture is shown in Fig. 3. We use ResNet as the feature extractor for its strong ability to learn features from labeled images. We have tried ResNet-50 and ResNet-101 in our experiments. Note that ResNet-50 has 16 bottleneck residual blocks and ResNet-101 has 33 bottleneck residual blocks. Taking 320×320 pixels RGB image as input, the feature extractor produces 2048 feature maps of size 10×10, as illustrated in Fig. 3.

The multi-grasp predictor has two components: classification layer and regression layer. To avoid overfitting, these two layers are convolutional layers instead of fully connected layers. The kernel size of these layers is 3×3. At each of the 10×10 locations of feature maps, the regression layer produces 5×$k$ offset output values for $k$ oriented anchor boxes. The parameterization of five offset values has been demonstrate in the left column of (2). At the same time, the classification layer produces 2×$k$ values (graspable score and ungraspable score of $k$ oriented anchor boxes) for each location on final feature map. Our model is one-stage detector, which directly predict grasps from the feature maps produced by feature extractor.

From another perspective, our classification layer predicts rough grasp rectangle centers, scales and orientations for grasps by determining whether the oriented anchor box centered at the given location is graspable. At the same time, regression layer refines grasp rectangle centers, scales and orientations. This is similar to [5], except that in [5] the refinement happens in the second stage.

*C. Matching Strategy*

During training, we assign a positive label to the oriented anchor box that satisfies the following conditions:

1) The centers of both oriented anchor box and the ground-truth rectangle locate in the same grid cell.

2) The orientation difference between oriented anchor box and ground-truth rectangle should below 90°/$k$.

The first condition is the point metric in [3], in which the center point of good grasp prediction is within a distance from at least one ground-truth grasp rectangle center. This also ensures that the center of the matched oriented anchor box is the nearest one to the ground-truth rectangle center.

The second condition guarantees that the matched oriented anchor box and ground-truth rectangle are within the same range of rotation angle. Compared with other oriented anchor boxes located at the same grid cell, the matched one has the least angle difference with ground-truth grasp rectangle. A larger $k$ makes the matched oriented anchor box more close to ground-truth grasp rectangle, which is helpful for regression loss in Section IV-D. However, a larger $k$ also introduces more examples that are negative and aggravates the imbalance between positive and negative examples, which brings difficulty for classification. From the thinking above, $k$ is set as 6 in our experiments.

To implement this matching strategy, we need to apply ceil function to the coordinates of the ground-truth rectangle center to find the grid cell where the matched oriented anchor box locate. Then we apply ceil function on the orientation of ground-truth rectangle to find the matched oriented box among the set of oriented anchor boxes at the same location.

Obviously, apply ceil function for three times is much easier than computing the Jaccard index, .i.e., the intersection over union, between an oriented rectangle and a rectangle parallel to horizontal axis. Previous works [4] [19] assign a reference rectangle (horizontal parallel rectangle) to be positive if its Jaccard index with ground-truth rectangle is above 0.5, which neglects the importance of angle. Our matching strategy take both location and rotation angle into account at a low computation cost. To be fair, we use Jaccard index for evaluation, as [4] [5] [10] [12] do. More detailed information about evaluation will be discussed in Section V-C.

### D. Loss Function

We adopt smooth $L_1$ [14] for regression loss. With the parameterization in (2), we define the regression loss as follow:

$$L_{reg}(\{t\}) = \sum_{i \in Positive} \sum_{m \in \{x,y,w,h,\theta\}} \text{smooth}_{L_1}(t_m^{(i)} - \hat{t}_m^{(i)}) \quad (3)$$

where $N$ is the number of positive oriented anchor boxes, which match with the ground-truth rectangles. $t_m$ is the offset predicted by the network. $\hat{t}_m$ is the corresponding ground-truth offset value. $t^{(i)}$ is a vector representing the 5 parametrized grasp offset values of the matched oriented anchor box. Parameterization is shown in (2). $t^{(i)}$ is selected from regression result in Fig. 3 by matching strategy.

Positive oriented anchor boxes are only small part of the entire oriented anchor boxes, the rest are negative. Due to this imbalance between positive and negative examples, the classification loss function will lead to no converge by back propagating all. We only propagate classification loss of positive examples and part of negative examples. Specifically, the number of negative examples is three times of the number of positive examples. We sort the unmatched oriented anchor boxes by their graspability score and select the top $3N$ boxes as negative examples. For classification, we adopt cross-entropy loss, which is defined as:

$$L_{cls}(\{p\}) = -\sum_{i \in Positive}^{N} \log(p_g^{(i)}) - \sum_{i \in Negative}^{3N} \log(p_u^{(i)}) \quad (4)$$

where $p_g^{(i)}$, the graspable output in Fig. 3, is the graspability score for the positive example. $p_u^{(i)}$, the ungraspable output in Fig. 3, is the ungraspability score for the negative example. $p^{(i)}$ is the vector of graspability score and ungraspability score.

Finally, our loss function is defined as:

$$L(\{p\},\{t\}) = \frac{1}{N}\left(L_{cls}(\{p\}) + \alpha L_{reg}(\{t\})\right) \quad (5)$$

The classification loss and regression loss are normalized with $N$ and balanced weight $\alpha$. In our work, $\alpha$ is set as 10.

## V. EXPERIMENTS SETUP

### A. Dataset

In order to compare with other algorithms, our models are trained and tested on Cornell Grasp Dataset. The dataset contains 885 images of 240 graspable objects. In each image, good grasps are labeled as positive grasp rectangles. For a given object, its grasp rectangles are varied in location, scale and orientation.

Like previous works, we divide the dataset into training set of 708 images and test set of 177 images in two different ways:

1) **Image-wise split** divides the images into training set and validation set at random. This aims to test the generalization ability of the network to new position and orientation of an object it has seen before.

2) **Object-wise split** divides the dataset at object instance level. All the images of an instance are put into the same set. This aims to test the generalization ability of the network to new object.

The training process for a deep neural network needs a large manually labeled dataset. While this kind of dataset is unavailable in most robotics applications. To apply deep learning in robotics, we need first to solve the problem of dataset. Researchers solve this problem from two aspects. First, pre-train the network in larger dataset [20] like ImageNet [21]. Second, expand the target dataset by data augmentation, which is also used in our work.

Compared with other datasets in deep learning, the Cornell Grasp Dataset is a small dataset. So extensive data augmentation is needed before feeding the data into the network. The data augmentation expands the dataset from different aspects. We take a center crop of 320x320 pixels with randomly translation up to 50 pixels in both x and y directions. This image patch is then randomly rotated up to 15 degrees in both clockwise and anti-clockwise direction. Then the image is randomly flipped horizontally or vertically. After that, we execute color and sharpness augmentation. Then we put the image into the network at the resolution of 320x320. Our augmentation is implemented online, which means every input image is a new image from pixel-level.

### B. Implementation Details

The feature extractors are convolutional layers pre-trained on RGB images of ImageNet, which helps the large convolutional neural network to avoid overfitting, especially when the dataset is limited.

Our models is implemented with Torch [22] for its great flexibility. For training and testing, our models run on a single NVIDIA TITAN-X (Pascal Architecture). For each of the models we tested, we employ the same training regimen. Limited by the graphic memory, the batch size is set as 16. Each model is trained end-to-end for 80k iterations. We use SGD with momentum of 0.9 to optimize our models. The learning rate is set as 0.0001 with a learning rate decay of 0.0001.

Table I  Accuracy under different Jaccard threshold, i.e.20%, 25%, 30%, 35%

| Approach | Setting | image-wise (%) | | | | object-wise (%) | | | |
|---|---|---|---|---|---|---|---|---|---|
| | | 20% | 25% | 30% | 35% | 20% | 25% | 30% | 35% |
| Guo et al. [4] | 1 scale and 1 aspect ratio | 93.8 | 93.2 | 91.0 | 85.3 | 85.1 | 82.8 | 79.3 | 74.1 |
| Guo et al. [4] | 3 scales and 3 aspect ratios | 88.1 | 86.4 | 83.6 | 76.8 | 90.8 | 89.1 | 85.1 | 80.5 |
| Chu et al. [5] | 3 scales and 3 aspect ratios | - | 96.0 | 94.9 | 92.1 | - | 96.1 | 92.7 | 87.6 |
| **Ours** | ResNet-50 | **98.87** | **97.74** | 94.92 | 89.83 | 95.48 | 94.91 | 89.26 | 84.18 |
| **Ours** | ResNet-101 | 98.31 | **97.74** | **96.61** | **95.48** | **97.74** | **96.61** | **93.78** | **91.53** |

Table II Accuracy under different angle threshold

| Rotation Angle Difference | image-wise | object-wise |
|---|---|---|
| | 25% | 25% |
| 30° | **97.74** | **96.61** |
| 25° | 97.74 | 96.04 |
| 20° | 97.18 | 95.48 |
| 15° | 94.35 | 93.22 |
| 10° | 86.44 | 85.31 |

## C. Evaluation

Similar to [4] [5] [10] [12], we also use rectangle metric to evaluate grasp detection results. In this metric, a predicted grasp is regarded as a good grasp if it satisfies both:

1) The rotation angle difference between predicted grasp and ground-truth grasp is within 30°.

2) The Jaccard index of the ground-truth grasp and predicted grasp is larger than 25%. The Jaccard index is defined as:

$$J(G,\hat{G}) = \frac{G \cap \hat{G}}{G \cup \hat{G}} \quad (6)$$

where $G$ is the area of predicted grasp rectangle and $\hat{G}$ is the area of ground-truth grasp rectangle. $G \cap \hat{G}$ is the intersection of these two rectangles. $G \cup \hat{G}$ is union of these two rectangles. Note that Jaccard index is only used in our evaluation, not in matching strategy.

## VI. RESULTS

For the images used to test the network, we just take a center crop of the images without any other augmentation. The images are then feed into the network one after another, not in the form of batch. We test the proposed models under different Jaccard index threshold. The detailed results of our methods and others methods are listed in Table I. The comparison between the methods of Guo et al. [4] demonstrates that 3 scales and 3 aspect ratios can largely improve the performance on object-wise split. This is maybe because the appearance of unknown object requires a thorough detection with multiple rectangle settings. The distribution of scale and aspect ratio for grasp rectangle between training set and test set of image-wise split is more similar than that of object-wise split. Therefore, algorithms need to tackle with new position, which places more emphasis on the orientation. Although ResNet-101 has the same accuracy as ResNet-50 on image-wise split under 25% Jaccard index, it outperforms ResNet-50 when the Jaccard index threshold is higher. Using ResNet-101 as feature extractor can be more robust to the change of Jaccard index threshold. The performance difference between ResNet-50 and ResNet-101 shows that employing deeper network is the another way to improve object-wise accuracy apart from multiple scales and aspect ratios. Our model based on ResNet-50 runs at the speed of 9.89 frames per second and our model based on ResNet-101 runs at the speed of 8.51 frames per second.

We have emphasized the importance of rotation angle in former sections. In Table II, our models are tested under different angle threshold. The performance of our model is still good when the threshold is bigger than 15°. Note that after threshold equal to or smaller than 15°, there is a sharper

Table III  Performance of different algorithms on Cornell Grasp Dataset.

| Approach | Algorithm | Accuracy (%) | |
|---|---|---|---|
| | | image-wise | object-wise |
| Jiang et al. [9] | Fast Search | 60.5 | 58.3 |
| Lenz et al. [3] | SAE, struct. reg. Two stage | 73.9 | 75.6 |
| Redmon et al. [10] | AlexNet, MultiGrasp | 88.0 | 87.1 |
| Kumra et al. [12] | ResNet-50×2, Multi-model Grasp Predictor | 89.21 | 88.96 |
| Guo et al. [4] | ZF-net, Hybrid network, 3 scales and 3 aspect ratios | 93.2* | 89.1 |
| Chu et al. [5] | VGG-16, Deep Grasp, 3 scales and 3 aspect ratios | 95.5 | 91.7 |
| | ResNet-50, Deep Grasp, 3 scales and 3 aspect ratios | 96.0 | 96.1 |
| Ours | ResNet-50, 6 default orientations | **97.74** | 94.92 |
| | ResNet-101, 6 default orientations | **97.74** | **96.61** |

Note: * signifies that this result is obtained under 1 scale and 1 aspect ratio.

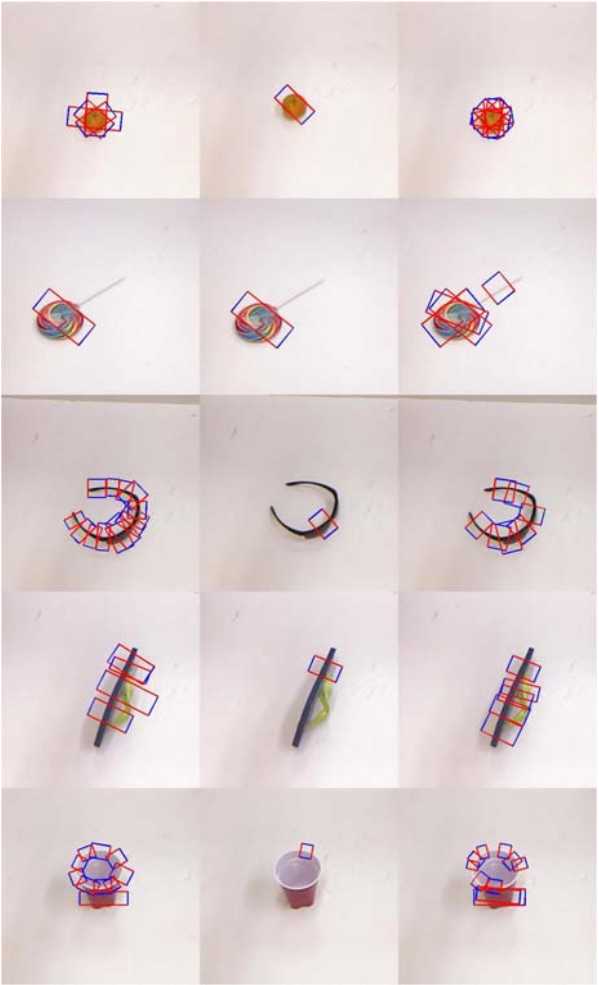

Fig. 4. Detection results in Cornell Grasp Dataset. The first column is ground-truth oriented rectangles. The second column is the visualization of Top1 grasp detection result. The third column is the results of multi grasps.

decline of accuracy, which may be relevant to the value of $k$ and the angle constrain in matching strategy.

Accuracy of the works on Cornell Grasp Dataset are listed in Table III. We can see that after VGG-16, deepening the network brings little improvement on the accuracy of image-wise split. In practice, we have 600 ($10 \times 10 \times k$, $k = 6$) predictions, which is one third of the prediction number in Chu et al. [5], preprinted on ArXiv in Feb 2018. We achieve better performance using a less dense prediction. Using much less predictions, our ResNet-50 model outperforms the state-of-the-art method [5] by 1.74% on image-wise split, which demonstrates that oriented anchor box mechanism provides a more accurate and efficient way for grasp detection. In view of the great improvement brought by multiple scales and aspect ratios on object-wise split accuracy, further improvement on our accuracy of object-wise split can be achieved by introducing multiple scales.

In Fig. 4, we visualize ground-truth rectangles and detection results of some objects in the test set of Cornell Grasp Dataset under image-wise splitting. The first column shows the ground-truth grasp rectangles of the objects. The second column visualizes the Top1 detection results of these object. Multi-grasp detection results are demonstrated in the third column. All the grasp rectangles in the third column have a graspable score over 0.5. The multi-grasp result of first row reveals the working mechanism of oriented anchor box, which predicts several rectangles at the same center but with different rotation angles. From the multi-grasp examples of the second line, we can see that our model predicts grasps from the shape of the objects rather than just fitting the annotation. Note that the grasp rectangles in multi-grasp detection results cover most grasp locations, even the one not labeled in ground-truth. Only a few predicted grasps have large overlap with others, which indicates our predicted grasps efficiently cover most of the representative grasp locations.

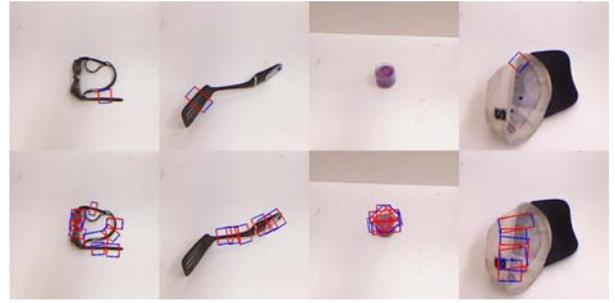

Fig. 5. Unsuccessful detection results. The first row is the detection result. The second row is the ground-truth oriented rectangle.

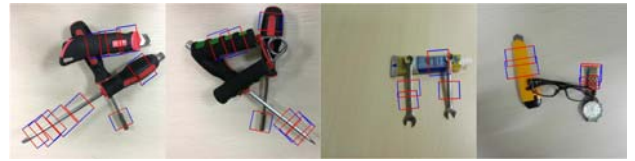

Fig. 6. Visualization of detection in more complex scene

Under the image-wise split accuracy of 97.74%, we has four unsuccessful detections in total. All these false detections results are shown in Fig. 5. The first row is the detection results and the second row is the ground-truth rectangles. Although same predictions do not satisfy the rectangle metric, these predictions are still feasible because the annotation is not exhaustive.

To test our model on more realistic and complex scenes, we test our ResNet-101 model, which is trained on image-wise split, with some pictures where the objects overlap with each other. The results is shown in Fig. 6. Some categories (knife, watch, wrench, wrist developer) never appear in Cornell Grasp Dataset. The glasses in Cornell dataset are usually dark glasses. The grasp rectangle for transparent glasses in the fourth column of Fig. 6 is more difficult to detect. Despite the occlusion, our model still has good performance under more realistic and complex scene. Besides, our model successfully predicts grasp for unseen objects.

## VII. CONCLUSION

We represent a new architecture for grasp detection with proposed oriented anchor box mechanism and new matching strategy. On Cornell Grasp Dataset, our best model outperforms current state-of-the-art model. Furthermore, our model can predict diverse grasps for an object.

Our future work will focus on detecting grasp locations for all the objects in an image and grasp relationship parse in the image where a pile of objects overlap with each other. A larger dataset with more detailed annotation will be collected. We will adapt the model to grasp diverse objects under a more realistic scene. The other future consideration is to speed up the algorithm by ameliorating the neural network like pruning its unnecessary channels.

ACKNOWLEDGMENT

This work was supported in part by NSFC No.91748208, the National Key Research and Development Program of China under grant No. 2017YFB1302200 and 2016YFB1000903, NSFC No. 61573268, and Shaanxi Key Laboratory of Intelligent Robots.

REFERENCES

[1] A. T. Miller and P. K. Allen, "Graspit! a versatile simulator for robotic grasping," Robotics & Automation Magazine, IEEE, vol. 11, no. 4, pp. 110–122, 2004.
[2] R. Pelossof, A. Miller, P. Allen, and T. Jebara, "An svm learning approach to robotic grasping," in IEEE International Conference on Robotics & Automation (ICRA), vol. 4. IEEE, 2004, pp. 3512–3518.
[3] I. Lenz, H. Lee, and A. Saxena, "Deep learning for detecting robotic grasps," The International Journal of Robotics Research, vol. 34, no. 4-5, pp. 705–724, 2015.
[4] D. Guo, F. Sun, H. Liu, T. Kong, B. Fang, and N. Xi, "A hybrid deep architecture for robotic grasp detection," in Proceedings of IEEE International Conference on Robotics and Automation, 2017, pp.1609–1614
[5] Chu, F. J., & Vela, P. A. (2018). Deep Grasp: Detection and Localization of Grasps with Deep Neural Networks. arXiv preprint arXiv:1802.00520.
[6] " Cornell grasping dataset," http://pr.cs.cornell.edu/grasping/rect data/data.php, 2013, accessed: 2017-09-01.
[7] K. He, X. Zhang, S. Ren, and J. Sun, "Deep residual learning for image recognition," in 2016 IEEE Conference on Computer Vision and Pattern Recognition (CVPR), pp. 770–778, June 2016.
[8] Saxena A, Driemeyer J, Ng A Y. Robotic grasping of novel objects using vision[J]. The International Journal of Robotics Research, 2008, 27(2): 157-173.
[9] Jiang Y, Moseson S, Saxena A. Efficient grasping from rgbd images: Learning using a new rectangle representation[C]//Robotics and Automation (ICRA), 2011 IEEE International Conference on. IEEE, 2011: 3304-3311
[10] J. Redmon and A. Angelova, "Real-time grasp detection using convolutional neural networks," in 2015 IEEE International Conference on Robotics and Automation (ICRA), pp. 1316–1322, May 2015.
[11] A. Krizhevsky, I. Sutskever, and G. E. Hinton, "Imagenet classification with deep convolutional neural networks," in Advances in Neural Information Processing Systems 25, pp. 1097–1105, 2012
[12] S. Kumra and C. Kanan, "Robotic grasp detection using deep convolutional neural networks," in Proceedings of the IEEE International Conference on Intelligent Robotic and Systems, 2017.
[13] Zeiler, M. D., & Fergus, R. (2014, September). Visualizing and understanding convolutional networks. In European conference on computer vision (pp. 818-833). Springer, Cham.
[14] Ren S, He K, Girshick R, Sun J. Faster r-cnn: Towards real-time object detection with region proposal networks. InAdvances in neural information processing systems 2015 (pp. 91-99).
[15] D. Guo, F. Sun, T. Kong, H. Liu, Deep vision networks for real-time robotic grasp detection, Int. J. Adv. Robot. Syst. 14 (1) (2017) 1–8.
[16] L. Pinto and A. Gupta, "Supersizing self-supervision: Learning to grasp from 50k tries and 700 robot hours," in Robotics and Automation (ICRA), 2016 IEEE International Conference on. IEEE, 2016, pp. 3406–3413
[17] Simonyan K, Zisserman A. Very deep convolutional networks for large-scale image recognition[J]. arXiv preprint arXiv:1409.1556, 2014.
[18] Huang J, Rathod V, Sun C, et al. Speed/accuracy trade-offs for modern convolutional object detectors[C]//IEEE CVPR. 2017.
[19] Liu W, Anguelov D, Erhan D, Szegedy C, Reed S, Fu CY, Berg AC. Ssd: Single shot multibox detector. InEuropean conference on computer vision 2016 Oct 8 (pp. 21-37). Springer, Cham.H. Poor, An Introduction to Signal Detection and Estimation. New York: Springer-Verlag, 1985, ch. 4.
[20] J. Yosinski, J. Clune, Y. Bengio, and H. Lipson, "How transferable are features in deep neural networks?," in Advances in neural information processing systems, pp. 3320–3328, 2014
[21] J. Deng, W. Dong, R. Socher, L.-J. Li, K. Li, and L. Fei-Fei,"ImageNet: A Large-Scale Hierarchical Image Database," in CVPR09,2009.
[22] Collobert R, Kavukcuoglu K, Farabet C. Torch7: A Matlab-like Environment for Machine Learning[C]// BigLearn, NIPS Workshop. 2011